\documentclass{article}

\usepackage{microtype}
\usepackage{graphicx}
\usepackage{subfigure}
\usepackage{booktabs} 
\usepackage{hyperref}
\usepackage{longtable}

\usepackage{pifont}

\usepackage[accepted]{icml2025}

\usepackage{amsmath}
\usepackage{amssymb}
\usepackage{mathtools}
\usepackage{amsthm}

\usepackage[capitalize,noabbrev]{cleveref}

\theoremstyle{plain}

\theoremstyle{definition}

\theoremstyle{remark}

\usepackage[textsize=tiny]{todonotes}
\usepackage[inline]{enumitem}
\usepackage{xspace}
\usepackage{float}

\newcommand{\ourmodel}{Real-TabPFN\xspace}

\icmltitlerunning{Real-TabPFN: Improving Tabular Foundation Models
via Continued Pre-training With Real-World Data}

\begin{document}

\twocolumn[
\icmltitle{Real-TabPFN: Improving Tabular Foundation Models \\
via Continued Pre-training With Real-World Data}



\icmlsetsymbol{equal}{*}

\begin{icmlauthorlist}
\icmlauthor{Anurag Garg}{1}
\icmlauthor{Muhammad Ali}{1}
\icmlauthor{Noah Hollmann}{2}
\icmlauthor{Lennart Purucker}{1}
\icmlauthor{Samuel Müller}{1,3}
\icmlauthor{Frank Hutter}{2,4,1}


\end{icmlauthorlist}

\icmlaffiliation{1}{Department of Informatik, University of Freiburg, Freiburg, Germany}
\icmlaffiliation{2}{Prior Labs GmbH, Freiburg, Germany}
\icmlaffiliation{3}{Meta, New York, USA (work done in at University of Freiburg)}
\icmlaffiliation{4}{ELLIS Institute Tübingen, Tübingen, Germany}

\icmlcorrespondingauthor{Anurag Garg}{garga@cs.uni-freiburg.de}
\icmlkeywords{Tabular Foundation Model, TabPFN}

\vskip 0.3in
]



\printAffiliationsAndNotice{}  

\begin{abstract}
Foundation models for tabular data, like TabPFN, achieve strong performance on small datasets when pre-trained solely on synthetic data. We show that this performance can be significantly boosted by a targeted continued pre-training phase. Specifically, we demonstrate that leveraging a small, curated collection of large, real-world datasets for continued pre-training yields superior downstream predictive accuracy compared to using broader, potentially noisier corpora like CommonCrawl or GitTables. Our resulting model, Real-TabPFN, achieves substantial performance gains on 29 datasets from the OpenML AutoML Benchmark.
\end{abstract}

\section{Introduction}
\label{introduction}

\begin{figure}[ht]
\vskip 0.2in
\begin{center}
\centerline{\includegraphics[width=0.9\columnwidth]{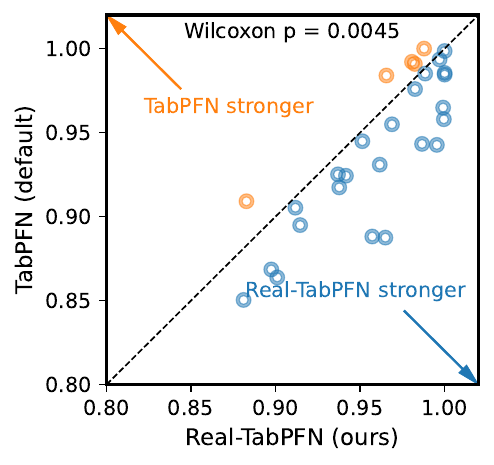}}
\caption{Per Dataset Normalized ROC Comparison of TabPFN (default) and \ourmodel (ours) on the 29 datasets from the OpenML AutoML Benchmark Datasets. Wilcoxon p refers to the two-sided Wilcoxon signed-rank test p value.}
\label{fig:per_ds_comparison}
\end{center}
\vskip -0.2in
\end{figure}







Until recently, traditional tree-based algorithms like XGBoost \cite{chen2016xgboost} and CatBoost \cite{prokhorenkova2018catboost} have consistently outperformed neural networks on tabular prediction tasks \cite{grinsztajn2022treebased}. However, TabPFNv2 has recently demonstrated improved performance on small datasets (up to 10,000 samples and 500 features), significantly advancing deep learning for tabular data.

Although TabPFNv2 delivers strong average performance, it is still not universally best-in-class: many datasets are better handled by well-tuned tree ensembles or by careful hyper-parameter searches on TabPFNv2 itself. 
It is not easy to improve the TabPFNv2 model further, since it has already been exhaustively trained on over 100 million synthetic tables that approximate a broad prior over tabular problems. However, even small accuracy gains could translate into tangible benefits, such as fewer hospital re-admissions or more precise credit-risk scoring, domains in which tabular data dominates.

We enhance TabPFNv2’s in-context learning by continuing to pre-train it on a carefully selected set of real-world tables from OpenML \cite{OpenML2013} and Kaggle\footnote{\url{https://www.kaggle.com}}. 
The resulting model, \textbf{\ourmodel}, consistently outperforms TabPFNv2 on the OpenML AutoMLBenchmark classification tasks (see~\cref{fig:per_ds_comparison}).  
In practice, \ourmodel serves as a stronger off-the-shelf baseline for tabular classification than the default TabPFNv2 model.

\textbf{Our contributions are:}
\begin{itemize}[label={\ding{227}}]

\item We empirically show that using real-world datasets with synthetic data during the pre-training of a tabular foundation model boosts in-context learning performance, opening a promising research direction.
\item The new model \textbf{\ourmodel} and its public weights\footnote{\href{https://huggingface.co/Prior-Labs/TabPFN-v2-clf/blob/main/tabpfn-v2-classifier-finetuned-zk73skhh.cpkt}{available on Hugging Face}}, an extension of TabPFNv2 obtained by continued pre-training on real-world data.
\ourmodel's in-context learning outperforms its predecessor on $29$ small tabular datasets.

\end{itemize}

\section{Related Work}

So far, tabular foundation models have been pre-trained solely on synthetic or real-world data. Our work aims to bridge this gap via continued pre-training.

\textbf{Synthetic‑only Tabular Foundation Models.}
TabPFN pre-trains a transformer \cite{Attention2017} on millions of synthetically generated tabular datasets, achieving strong in-context learning performance on small datasets~\cite{hollmann2023tabpfn,
hollmann2025tabpfn}. Several extensions retain the synthetic‑data recipe: TabForestPFN~\cite{denBreejen2024tabforestpfn} augments TabPFN with more complex, decision‑boundary‑oriented generators, and TabICL~\citep{qu2025tabicl} scales to tables with 500k rows.

\textbf{Real‑data Foundation Models.} In parallel, purely real‑data approaches emerged. TabDPT \cite{ma2024tabdpt} couples retrieval-based self‑supervision with discriminative transformers and is trained on real-world data collected from OpenML. TabuLa‑8B~\cite{schmidt2024tabula} adapts an Llama 3-8B backbone via language modeling over serialized tables, demonstrating that large LLMs can transfer to tabular few-shot prediction after real‑world pre-training.

\section{Pre-training and Evaluation Data}

To study continued pre-training, we had to decide on the data for continuing pre-training and for evaluation. 

\textbf{Evaluation Data.} We adopt the same datasets as used by \citet{hollmann2025tabpfn} for TabPFNv2 to evaluate our method. 
We use the same 29 datasets from the OpenML AutoML Benchmark \cite{gijsbers2023amlbautomlbenchmark}; see Appendix \ref{appendix:evaluation-datasets}. All datasets contain up to 10,000 samples and 500 features.

\textbf{Continued Pre-training Data.}
Unlike the domains of natural language processing and computer vision, where many carefully curated datasets, such as ImageNet~\cite{deng2009imagenet}, COCO~\cite{lin2014microsoft}, FineWeb~\cite{penedo2024fineweb}, and C4~\cite{raffel2020exploring} are available, comparably high-quality resources for tabular learning remain scarce.

\begin{figure}[ht]
\begin{center}
\centerline{\includegraphics[width=\columnwidth]{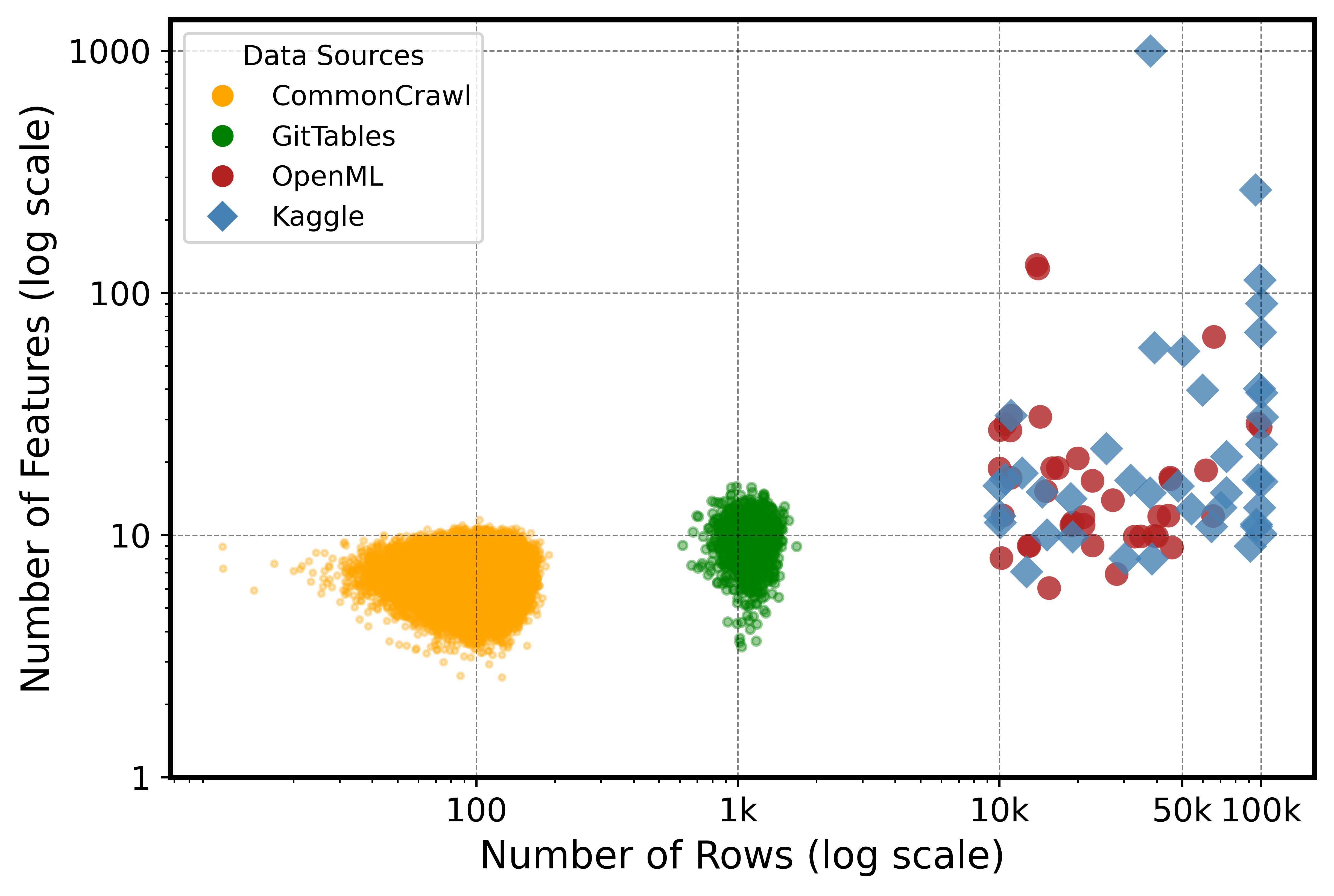}} 
\caption{Distribution of dataset sizes (number of rows and features) from various sources. The prevalence of smaller datasets in broad corpora like CommonCrawl and GitTable contrasts with the larger datasets from OpenML and Kaggle. }
\label{fig:dataset_size_distribution}
\end{center}
\end{figure}

Recent efforts have attempted to address this gap.  Notable contributions include the Web Table Corpus \cite{madata208}, TabLib \cite{tablibdataset627mtables}, and GitTables \cite{GitTables}. Beyond these, researchers either generate synthetic datasets or rely on existing repositories like UCI \cite{UCI_Repo}, OpenML \cite{OpenML2013}, and Kaggle. We adopt the latter, manually curating $71$ high-quality datasets from OpenML and Kaggle. \cref{fig:dataset_size_distribution} shows the distribution of the number of features and rows across datasets from different sources; Appendix \ref{appx.Training-Datasets} lists the curated datasets used for continued pre-training.
We apply minimal preprocessing to the $71$ datasets: categorical features are encoded with Scikit-learn’s \cite{scikit-learn} OrdinalEncoder; 
and if the target variable has more than ten classes, we retain the nine most common classes and merge the remainder into a single tenth class.

\paragraph{Data Contamination.}
\label{data_contamination}
We carefully avoided data contamination \cite{jiang2024investigatingdatacontaminationpretraining} between training and the evaluation set. We implemented a multi-tiered filtering process: 
\textbf{(1)} We only select datasets exceeding 10,000 samples since all our evaluation datasets have fewer samples.
\textbf{(2)} We cross-referenced dataset IDs, names, and shapes to identify potential duplicates.
\textbf{(3)} We compared feature names across datasets to detect similar or identical data structures.
\textbf{(4)} We generated hashes of both rows and columns to identify potential data duplications at a granular level.
\textbf{(5)} We manually inspected metadata to ensure no training and evaluation datasets share a common source dataset or a subsampled version of that dataset.
\\
We exclude any dataset from the pre-training data that does not meet these criteria. 

\section{Method: Continued Pre-training of TabPFN}
\label{method:section}
Our method bridges purely synthetic training (e.g., TabPFN~\cite{hollmann2025tabpfn,hollmann2023tabpfn}) and purely real-data training (e.g., TabDPT~\cite{ma2024tabdpt}) by leveraging the complementary strengths of both paradigms.

Concretely, we adopt a \emph{two-stage approach}. 
\textbf{Stage~1} relies on the original TabPFNv2 checkpoint, pre-trained by \citet{hollmann2025tabpfn} on a large, diverse set of synthetic tables. This serves as our starting point.
\textbf{Stage 2} continues pre-training exclusively on a curated collection of heterogeneous real-world tables.
This approach contrasts with \emph{mixed} training, where synthetic and real samples are fed to the model simultaneously, as in \citet{d2025synthetictabulardatageneration}. We opted for a two-stage approach as it is easier to apply and builds directly on a strong existing synthetic base model.

Although continued pre-training has shown remarkable success in language models \cite{gururangan-etal-2020-dont}, its potential for tabular foundation models remains largely unexplored. By pre-training on diverse real datasets rather than narrow task-specific data, our approach improves generalization while preserving cross-domain adaptability.

To enable robust continued pre-training, we retained the original TabPFNv2 architecture and trained with a reduced learning rate of $3 \times 10^{-7}$ using the AdamW optimizer \cite{loshchilov2017decoupled} together with a linear warm‑up followed by a cosine annealing schedule \cite{loshchilov2017sgdr}.

Moreover, we added a regularizer penalizing distance to the L2-Starting-Point (L2-SP) \citep{li2018explicitinductivebiastransfer} to the pre-training objective. This 
penalizes large deviations from the initial pre-trained weights, and is used to mitigate catastrophic forgetting \citep{Kirkpatrick_2017}. 
More precisely, let $\mathbf{w}^{0}$ denote the parameter vector of the pre-trained base model from which continued pre-training begins. 
The L2-SP penalty then regularizes the model parameters towards this initial vector. It is formally defined as:
\begin{equation*}
    \label{eq:l2-sp}
\Omega(\mathbf{w}) = \frac{\alpha}{2} \left\lVert \mathbf{w} - \mathbf{w}^{0} \right\rVert_{2}^{2} ,
\end{equation*}
where $\alpha$ controls the strength of the regularization penalty, and $\left\lVert \cdot \right\rVert_{2}$ denotes the L2 norm. We add the L2 norm to the cross-entropy loss: 
$\mathcal{L} \;=\; \mathcal{L}_{\text{CE}} \;+\; \Omega(\mathbf{w})$ to obtain our final pre-training objective. 
We used a regularization strength $\alpha$ of 0.003. 

We continued pre-training for 20,000 steps with a batch size of 1 (i.e., a single dataset). Choosing a batch size of 1 is a simple approach that naturally handles the varying feature dimensions of real-world datasets without requiring padding or truncation. Critically, it also allowed us to maximize the training context for each dataset up to 20,000 samples, limited primarily by GPU memory, rather than by batching constraints.
For datasets larger than this limit, we sample uniformly up to 20,000 rows.
Per batch, we split the data into 60\% context (TabPFN's training data) and 40\% query (TabPFN's testing data) for the forward pass.
The training was performed on a single Nvidia RTX 2080 Ti GPU. To stay within GPU memory limits, we further capped each dataset at 400,000 total cells, adjusting the number of samples accordingly for datasets with too many attributes.

\section{Experiments and Results}

We follow the evaluation protocol of \citet{hollmann2025tabpfn} and evaluate \ourmodel with 10-fold cross-validation per dataset. 
Furthermore, we reuse the performance values for additional baselines from the results reported by \citet{hollmann2025tabpfn}.
The baselines were tuned for ROC-AUC via five-fold cross-validated random search under a four-hour time budget.
We run \ourmodel and TabPFNv2 ourselves without hyperparameter tuning to focus on their in-context learning performance. 

\begin{figure}[t]
    \begin{center}
    \centerline{\includegraphics[width=0.8\columnwidth]{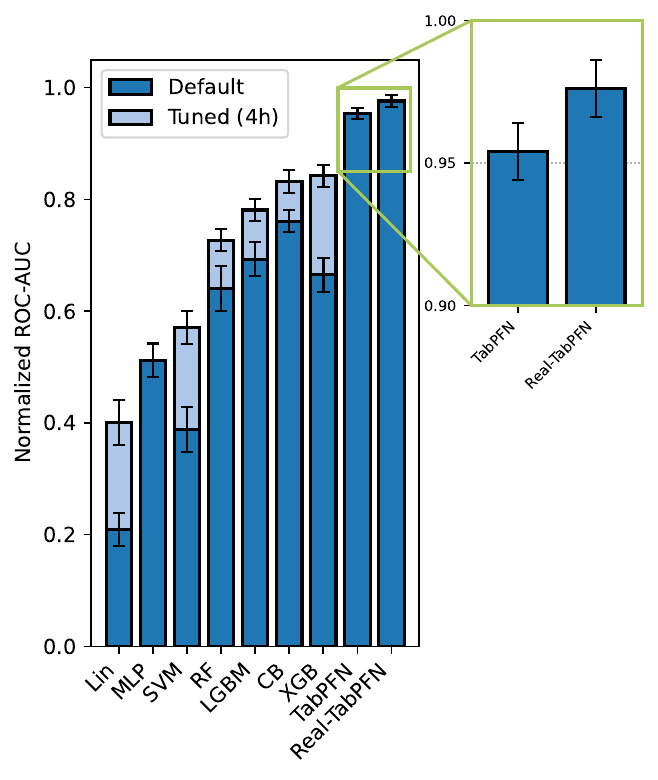}}
    \caption{Mean Normalized ROC AUC Comparsion of \ourmodel with all the default and the tuned versions of the baselines on the AutoMLBenchmark. Scores were normalized per dataset, with 1.0 representing the best and 0.0 the worst performance with respect to all baselines. }
    \label{fig:method-roc-comparison}
    \end{center}
    \vskip -0.2in
\end{figure}

\cref{fig:per_ds_comparison} compares TabPFNv2 and \ourmodel per dataset. We observe that \ourmodel significantly outperforms TabPFNv2. 
Additionally, \cref{fig:method-roc-comparison} shows that \ourmodel improves the mean normalized ROC-AUC from 0.954 to 0.976 and naturally  outperforms all baselines on average, like TabPFNv2.
We provide a table with various additional performance metrics for all methods in \cref{appendix:main results}.

\begin{figure}[ht]
    \begin{center}
    \centerline{\includegraphics[width=0.8\columnwidth]{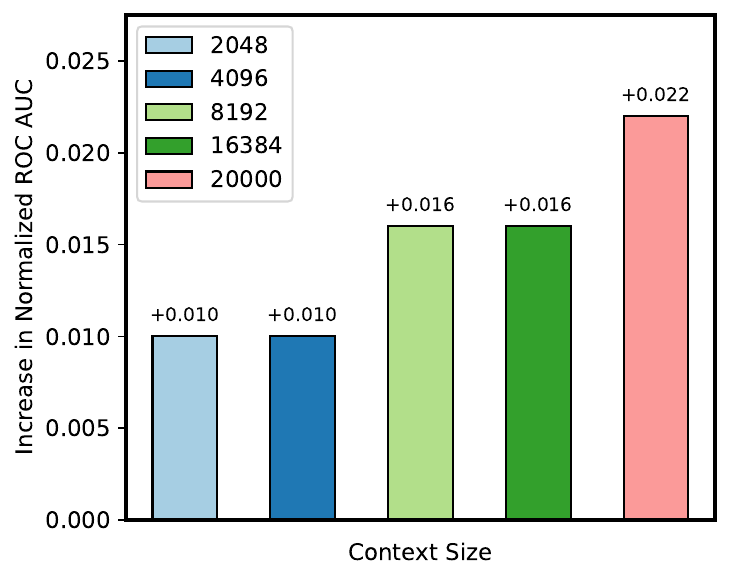}}
    \caption{Increase in normalized ROC AUC as the continued-pre-training context grows. The gains are shown relative to the base TabPFNv2 model performance which was synthetically pre-trained with 2,048 context size.}
    \label{fig:seq-len-exp}
    \end{center}
    \vskip -0.2in
\end{figure}

\begin{figure}[ht]
    \begin{center}
    \centerline{\includegraphics[width=0.8\columnwidth]{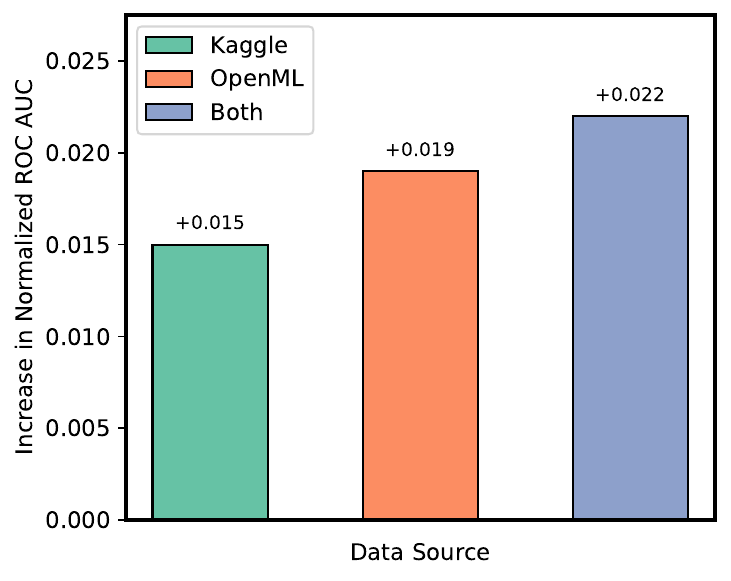}}
    \caption{Increase in normalized ROC AUC as the training data source is varied. The gains are shown relative to the base TabPFNv2 model performance which was synthetically pre-trained.}
    \label{fig:data-source-exp}
    \end{center}
    \vskip -0.2in
\end{figure}

\begin{figure}[ht]
    \begin{center}
    \centerline{\includegraphics[width=0.8\columnwidth]{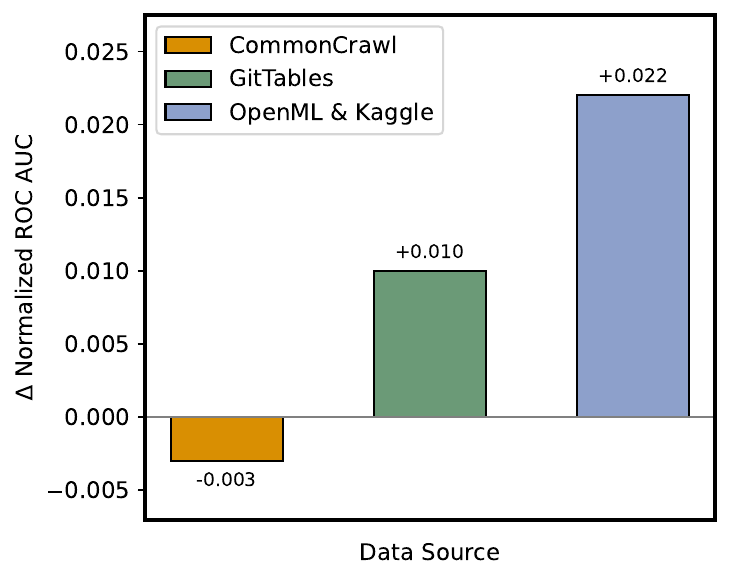}}
    \caption{Change in normalized ROC AUC as the training data source is varied. The changes are shown relative to the base TabPFNv2 model performance which was synthetically pre-trained.}
    \label{fig:git_vs_cc}
    \end{center}
    \vskip -0.2in
\end{figure}

\paragraph{Effect of Context Size.} 
We investigate the impact of the size of datasets during continued pre-training by testing pre-training with datasets from $2{,}048$ to $20{,}000$ (our GPU memory limit) samples. Figure~\ref{fig:seq-len-exp} shows that downstream accuracy increases with a larger context size.

\paragraph{OpenML vs Kaggle.}  To understand the impact of our final curated $71$ training data on model performance, we repeated continued pre-training with three corpora: (1) only \textsc{Kaggle}, (2) only \textsc{OpenML}, and (3) the union of both. As Figure~\ref{fig:data-source-exp} shows, \textsc{OpenML} alone delivers a \mbox{$+0.019$} gain, while \textsc{Kaggle} alone gives \mbox{$+0.015$}.  Combining them yields the strongest boost, \mbox{$+0.022$}, confirming that heterogeneous sources provide complementary supervision signals. This finding indicates that while OpenML datasets provide slightly better performance individually, the combination of both data sources yields the best performing model.

\paragraph{Effect of Training Data Source.} To evaluate the impact of different training data sources, we also experimented with two alternative corpora: (1) \textsc{CommonCrawl} ~\cite{yin2020tabertpretrainingjointunderstanding} and (2) \textsc{GitTables} ~\cite{tran2024tabularfmopenframeworktabular}. We applied aggressive filtering by evaluating datasets with Logistic Regression \cite{cox1958logistic} and Random Forest \cite{randomforest} and subsequently removing noisy datasets, followed by our data contamination pipeline (see \cref{data_contamination}). This resulted in approximately $97,000$ CommonCrawl and $658$ GitTables datasets.

\cref{fig:git_vs_cc} compares performance using these two corpora. The model trained on CommonCrawl (approximately 100 data points and 7 features on average per dataset; see \cref{fig:dataset_size_distribution}) exhibits decreased performance, primarily because the small dataset size did not sufficiently benefit the model during the continued pre-training phase, ultimately leading to a performance drop. 

In contrast, GitTables (approximately 1000 data points and 9 features on average per dataset; see \cref{fig:dataset_size_distribution}) leads to performance improvements. The biggest performance improvements are achieved with our manually curated OpenML and Kaggle datasets (10k to 100k data points and on average tens of features). We intentionally chose a smaller, curated set of datasets from OpenML and Kaggle to effectively prevent data contamination, which is why we did not combine them with GitTables or CommonCrawl datasets.

\section{Conclusion and Future Work}

We show that continued pre-training of TabPFNv2 on curated, real-world tabular data yields a stronger default model, \ourmodel, which we will open-source. Bridging the synthetic-to-real gap, \ourmodel outperforms the default TabPFNv2 on most of the datasets and outperforms every other state-of-the-art baseline on all evaluated datasets. Additional experiments deliver the same message: seeing \emph{more context}—whether temporal (longer windows) or statistical (a richer mix of bigger datasets) during continued pre-training produces larger improvements. 



\section*{Acknowledgements}
This research was funded by the Deutsche Forschungsgemeinschaft (DFG, German Research Foundation) under grant number 539134284, through EFRE (FEIH\_2698644) and the state of Baden-Württemberg. We also acknowledge funding by the European Union (via ERC Consolidator Grant DeepLearning 2.0, grant no.~101045765). Views and opinions expressed are however those of the author(s) only and do not necessarily reflect those of the European Union or the European Research Council. Neither the European Union nor the granting authority can be held responsible for them. L.P. acknowledges funding by the Deutsche Forschungsgemeinschaft (DFG, German Research Foundation) under SFB 1597 (SmallData), grant number 499552394; and funding by ELSA – European Lighthouse on Secure and Safe AI funded by the European Union under Grant Agreement No. 101070617.
F.H. acknowledges the financial support of the Hector Foundation. Finally, we thank the reviewers for their feedback, which has helped us improve the manuscript. 

\begin{center}
\includegraphics[width=0.2\textwidth]{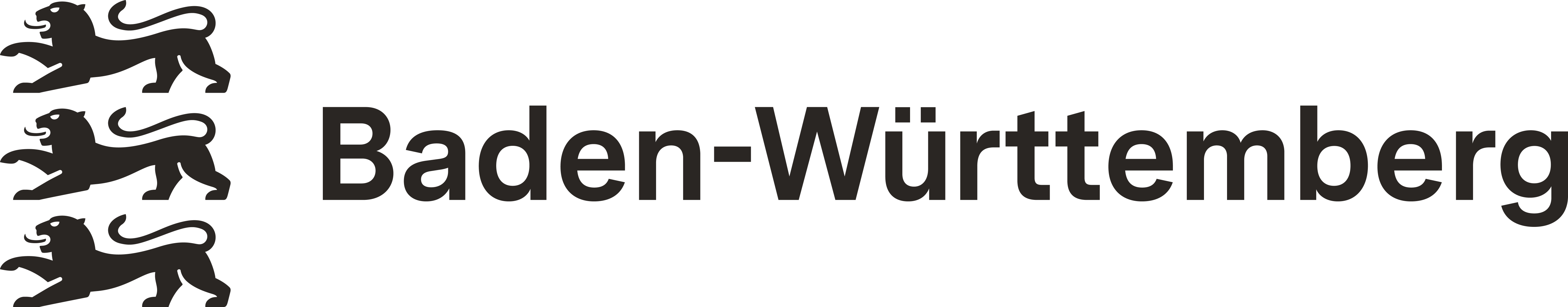} ~~~ \includegraphics[width=0.2\textwidth]{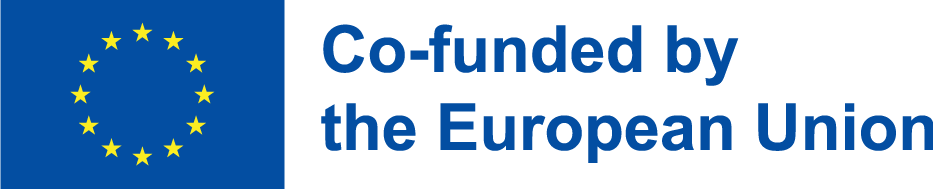}
\end{center}
\newpage

\bibliography{example_paper}
\bibliographystyle{icml2025}

\newpage
\appendix
\onecolumn

\section{Training Datasets}
\label{appx.Training-Datasets}
The following table lists the $71$ datasets curated for continued pre-training, along with their source and access link. 

\begin{longtable}{p{10cm}p{4cm}}
\toprule
\textbf{Name} & \textbf{Source} \\
\midrule
\endfirsthead
\toprule
\textbf{Name} & \textbf{Source} \\
\midrule
\endhead

\href{https://www.kaggle.com/datasets/himselfthedecker/aam-avaliacao-dataset}{aam\_avaliacao\_dataset} & Kaggle \\
\href{https://www.kaggle.com/datasets/rohanshetty678/air-traffic-data}{Air Traffic Data} & Kaggle \\
\href{https://www.kaggle.com/datasets/stefadp/ansibledefectsprediction}{ansible-defects-prediction} & Kaggle \\
\href{https://www.kaggle.com/datasets/nehaprabhavalkar/av-healthcare-analytics-ii}{AV Healthcare Analytics II} & Kaggle \\
\href{https://www.kaggle.com/datasets/tarunchilkur/client}{Candidate Selection} & Kaggle \\
\href{https://www.kaggle.com/datasets/sulianova/cardiovascular-disease-dataset}{Cardio Disease} & Kaggle \\
\href{https://www.kaggle.com/datasets/mlg-ulb/creditcardfraud}{CC\_Fraud\_Dataset} & Kaggle \\
\href{https://www.kaggle.com/datasets/aniketng21600/crop-damage-information-in-india}{Classification - Crop Damages in India (2015-2019)} & Kaggle \\
\href{https://www.kaggle.com/datasets/christianlillelund/csgo-round-winner-classification}{CSGO Round Winner Classification} & Kaggle \\
\href{https://www.kaggle.com/datasets/vpkprasanna/flower-type-prediction-machine-hack}{Flower Type Prediction Machine Hack} & Kaggle \\
\href{https://www.kaggle.com/datasets/gunner38/horseracing/data}{Horse Racing - Tipster Bets} & Kaggle \\
\href{https://www.kaggle.com/datasets/kanuriviveknag/road-accidents-severity-dataset}{How severe the accident could be} & Kaggle \\
\href{https://www.kaggle.com/datasets/shivan118/hranalysis}{HR Analysis Case Study} & Kaggle \\
\href{https://www.kaggle.com/datasets/anshika2301/hr-analytics-dataset}{HR analysis} & Kaggle \\
\href{https://www.kaggle.com/datasets/pankeshpatel/hrcommasep}{hr-comma-sep} & Kaggle \\
\href{https://www.kaggle.com/datasets/jsrojas/ip-network-traffic-flows-labeled-with-87-apps}{ip-network-traffic-flows-labeled-with-87-apps} & Kaggle \\
\href{https://www.kaggle.com/datasets/pawan2905/jantahack-cross-sell-prediction}{Janatahack cross-sell prediction} & Kaggle \\
\href{https://www.kaggle.com/code/neerunaveenjakhar/janatahack-machine-learning-for-banking/data}{JanataHack Machine Learning for Banking} & Kaggle \\
\href{https://www.kaggle.com/datasets/mamtadhaker/lt-vehicle-loan-default-prediction}{L\&T Vehicle Loan Default Prediction} & Kaggle \\
\href{https://www.kaggle.com/datasets/benfattori/league-of-legends-diamond-games-first-15-minutes}{League of Legends Diamond Games (First 15 Minutes)} & Kaggle \\
\href{https://www.kaggle.com/datasets/ppsheth91/two-target-variables-classification-problem}{Multiple target variable classification - Hackathon} & Kaggle \\
\href{https://www.kaggle.com/datasets/btphan/online-news-popularity-dataset}{Online News Popularity} & Kaggle \\
\href{https://www.kaggle.com/datasets/henrysue/online-shoppers-intention}{Online Shopper's Intention} & Kaggle \\
\href{https://www.kaggle.com/datasets/eswarchandt/phishing-website-detector}{Phishing website Detector} & Kaggle \\
\href{https://www.kaggle.com/datasets/shashwatwork/phishing-dataset-for-machine-learning}{Phishing websites Data} & Kaggle \\
\href{https://www.kaggle.com/datasets/dylanli/pump-it-up-data-mining-the-water-table?select=Pump_it_Up_Data_Mining_the_Water_Table_-_Training_set_values.csv}{Pump it Up Data Mining the Water Table} & Kaggle \\
\href{https://www.kaggle.com/code/franciscoescobar/richter-s-predictor-modeling-earthquake-damage}{Richter's Predictor Modeling Earthquake Damage} & Kaggle \\
\href{https://www.kaggle.com/datasets/kartikjaspal/server-logs-suspicious}{Server Logs - Suspicious} & Kaggle \\
\href{https://www.kaggle.com/datasets/lucidlenn/sloan-digital-sky-survey}{Sloan Digital Sky Survey DR14} & Kaggle \\
\href{https://www.kaggle.com/datasets/muhakabartay/sloan-digital-sky-survey-dr16}{Sloan Digital Sky Survey DR16} & Kaggle \\
\href{https://www.kaggle.com/datasets/raosuny/success-of-bank-telemarketing-data}{Success of Bank Telemarketing Data} & Kaggle \\
\href{https://www.kaggle.com/datasets/brajeshmohapatra/term-deposit-prediction-data-set}{Term Deposit Prediction Data Set} & Kaggle \\
\href{https://www.kaggle.com/datasets/danielamigo/trajectorybasedshipclassification/data}{trajectory-based-ship-classification} & Kaggle \\
\href{https://www.kaggle.com/datasets/mhdzahier/travel-insurance}{Travel Insurance} & Kaggle \\
\href{https://openml.org/search?type=data&status=active&id=4135}{Amazon\_employee\_access} & OpenML \\
\href{https://www.openml.org/search?type=data&sort=runs&status=active&id=1459}{artificial-characters} & OpenML \\
\href{https://openml.org/search?type=data&status=active&id=44234}{Bank\_marketing\_data\_set\_UCI} & OpenML \\
\href{https://www.openml.org/search?type=data&status=active&id=251}{BNG(breast-w)} & OpenML \\
\href{https://www.openml.org/search?type=data&status=active&id=137}{BNG(tic-tac-toe)} & OpenML \\
\href{https://www.openml.org/search?type=data&sort=runs&status=active&id=41434}{Click\_prediction\_small} & OpenML \\
\href{https://openml.org/search?type=data&status=active&id=4154}{CreditCardSubset} & OpenML \\
\href{https://www.openml.org/d/40668}{connect\_4} & OpenML \\
\href{https://www.openml.org/search?type=data&sort=runs&status=active&id=1471}{eeg-eye-state} & OpenML \\
\href{https://openml.org/search?type=data&status=active&id=151}{electricity} & OpenML \\
\href{https://www.openml.org/search?type=data&status=active&id=846&sort=runs}{elevators} & OpenML \\
\href{https://openml.org/search?type=data&status=active&id=43551}{Employee-Turnover-at-TECHCO} & OpenML \\
\href{https://openml.org/search?type=data&status=active&id=1044}{eye\_movements} & OpenML \\
\href{https://www.openml.org/search?type=data&status=active&id=41787&sort=runs}{FOREX\_eurpln-hour-High} & OpenML \\
\href{https://www.openml.org/search?type=data&sort=runs&status=active&id=1477}{gas-drift-different-concentrations} & OpenML \\
\href{https://www.openml.org/search?type=data&sort=runs&status=active&id=1476}{gas-drift} & OpenML \\
\href{https://openml.org/search?type=data&status=active&id=23512}{higgs} & OpenML \\
\href{https://www.openml.org/search?type=data&status=active&id=821&sort=runs}{house\_16H} & OpenML \\
\href{https://www.openml.org/search?type=data&status=active&id=843&sort=runs}{house\_8L} & OpenML \\
\href{https://openml.org/search?type=data&status=active&id=44201}{Intersectional-Bias-Assessment-(Training-Data)} & OpenML \\
\href{https://openml.org/search?type=data&status=active&id=43904}{law-school-admission-binary} & OpenML \\
\href{https://www.openml.org/search?type=data&sort=runs&status=active&id=40679}{magic} & OpenML \\
\href{https://openml.org/search?type=data&status=active&id=1120}{MagicTelescope} & OpenML \\
\href{https://openml.org/search?type=data&status=active&id=43617}{Medical-Appointment} & OpenML \\
\href{https://www.openml.org/search?type=data&status=active&id=41671&sort=runs}{microaggregation2} & OpenML \\
\href{https://www.openml.org/search?type=data&sort=runs&id=901&status=active}{fried} & OpenML \\
\href{https://openml.org/search?type=data&status=active&id=1046}{mozilla4} & OpenML \\
\href{https://www.openml.org/search?type=data&status=active&id=43923&sort=runs}{mushroom} & OpenML \\
\href{https://openml.org/search?type=data&status=active&id=44226}{NewspaperChurn} & OpenML \\
\href{https://openml.org/search?type=data&status=active&id=1568}{nursery} & OpenML \\
\href{https://www.openml.org/search?type=data&status=active&id=45067&sort=runs}{okcupid\_stem} & OpenML \\
\href{https://www.openml.org/search?type=data&status=active&id=32&sort=runs}{pendigits} & OpenML \\
\href{https://openml.org/search?type=data&status=active&id=4534&sort=runs}{PhishingWebsites} & OpenML \\
\href{https://www.openml.org/search?type=data&sort=runs&status=active&id=44122}{pol} & OpenML \\
\href{https://www.openml.org/search?type=data&status=active&id=46676&sort=runs}{WBCAtt} & OpenML \\
\href{https://www.openml.org/search?type=data&sort=runs&id=1461&status=active}{Bank Marketing} & OpenML \\
\href{https://www.openml.org/search?type=data&sort=runs&id=43039&status=active}{Internet Firewall Data} & OpenML \\

\bottomrule
\end{longtable}

\section{Evaluation Datasets}
\label{appendix:evaluation-datasets}
We use the same evaluation suite as TabPFNv2 to ensure direct comparability of results. All classification tasks from the AutoML Benchmark with fewer 10,000 samples and 500 features. The benchmark comprises diverse real-world tabular datasets, curated for complexity, relevance, and domain diversity.

\renewcommand{\arraystretch}{1.5}
{\small
\begin{longtable}{p{3cm} p{2cm} p{3cm} p{1.2cm} p{1.5cm} p{1.5cm} p{2.5cm}}
\toprule
\textbf{Name} & \textbf{OpenML ID} & \textbf{Domain} & \textbf{Features} & \textbf{Samples} & \textbf{Targets} & \textbf{Categorical Feats.} \\
\midrule
\endhead

\bottomrule
\multicolumn{7}{p{\textwidth}}{} \\
\endfoot

ada & 41156 & Census & 48 & 4147 & 2 & 0 \\
Australian & 40981 & Finance & 14 & 690 & 2 & 8 \\
blood-transfusion-service-center & 1464 & Healthcare & 4 & 748 & 2 & 0 \\
car & 40975 & Automotive & 6 & 1728 & 4 & 6 \\
churn & 40701 & Telecommunication & 20 & 5000 & 2 & 4 \\
cmc & 23 & Public Health & 9 & 1473 & 3 & 7 \\
credit-g & 31 & Finance & 20 & 1000 & 2 & 13 \\
dna & 40670 & Biology & 180 & 3186 & 3 & 180 \\
eucalyptus & 188 & Agriculture & 19 & 736 & 5 & 5 \\
first-order-theorem-proving & 1475 & Computational Logic & 51 & 6118 & 6 & 0 \\
GesturePhase Segmentation Processed & 4538 & Human-Computer Interaction & 32 & 9873 & 5 & 0 \\
jasmine & 41143 & Natural Language Processing & 144 & 2984 & 2 & 136 \\
kc1 & 1067 & Software Engineering & 21 & 2109 & 2 & 0 \\
kr-vs-kp & 3 & Game Strategy & 36 & 3196 & 2 & 36 \\
madeline & 41144 & Artificial & 259 & 3140 & 2 & 0 \\
mfeat-factors & 12 & Handwriting Recognition & 216 & 2000 & 10 & 0 \\
ozone-level-8hr & 1487 & Environmental & 72 & 2534 & 2 & 0 \\
pc4 & 1049 & Software Engineering & 37 & 1458 & 2 & 0 \\
philippine & 41145 & Bioinformatics & 308 & 5832 & 2 & 0 \\
phoneme & 1489 & Audio & 5 & 5404 & 2 & 0 \\
qsar-biodeg & 1494 & Environmental & 41 & 1055 & 2 & 0 \\
Satellite & 40900 & Environmental Science & 36 & 5100 & 2 & 0 \\
segment & 40984 & Computer Vision & 16 & 2310 & 7 & 0 \\
steel-plates-fault & 40982 & Industrial & 27 & 1941 & 7 & 0 \\
sylvine & 41146 & Environmental Science & 20 & 5124 & 2 & 0 \\
vehicle & 54 & Image Classification & 18 & 846 & 4 & 0 \\
wilt & 40983 & Environmental & 5 & 4839 & 2 & 0 \\
wine-quality-white & 40498 & Food and Beverage & 11 & 4898 & 7 & 0 \\
yeast & 181 & Biology & 8 & 1484 & 10 & 0 \\
\end{longtable}
}

\newpage
\section{Performance Comparison on 29 AMLB Classification Datasets}
\label{appendix:main results}
Scores are normalized on all the baselines (0 = worst, 1 = best) per dataset; all
methods are tuned for ROC-AUC, so secondary metrics may not
reflect their true rank.
\begingroup
  \setlength{\tabcolsep}{5pt}     
  \renewcommand{\arraystretch}{1.3} 

  \makebox[\textwidth][c]{%
  \resizebox{\textwidth}{!}{%
  \begin{tabular}{l|ccccc|ccccc|c}
  \hline
   & \multicolumn{5}{c|}{Mean Normalized} & \multicolumn{5}{c|}{Mean} & Mean \\
   & ROC & Acc. & F1 & CE & ECE & ROC & Acc. & F1 & CE & ECE & Time (s) \\
   & ($\uparrow$) & ($\uparrow$) & ($\uparrow$) & ($\downarrow$) & ($\downarrow$) &
     ($\uparrow$) & ($\uparrow$) & ($\uparrow$) & ($\downarrow$) & ($\downarrow$) & \\
  \hline
  \ourmodel        & \textbf{0.976} & \textbf{0.932} & \textbf{0.939} & \textbf{0.011} & \textbf{0.107} & \textbf{0.932} & \textbf{0.862} & \textbf{0.771} & 0.337 & \textbf{0.040} & 2.921 \\
                   & \textbf{$\pm$0.01} & \textbf{$\pm$0.01} & \textbf{$\pm$0.01} & \textbf{$\pm$0.00} & \textbf{$\pm$0.01} & \textbf{$\pm$0.01} & \textbf{$\pm$0.02} & \textbf{$\pm$0.04} & $\pm$0.03 & \textbf{$\pm$0.01} & $\pm$0.57 \\
  \hline
  TabPFN           & 0.954 & 0.906 & 0.920 & 0.036 & 0.111 & 0.929 & 0.857 & 0.767 & 0.347 & 0.042 & 2.793 \\
  (default)        & $\pm$0.01 & $\pm$0.01 & $\pm$0.01 & $\pm$0.01 & $\pm$0.02 & $\pm$0.01 & $\pm$0.02 & $\pm$0.04 & $\pm$0.03 & $\pm$0.01 & $\pm$0.49 \\
  \hline
  Autogluon(V1,    & 0.928 & 0.888 & 0.916 & 0.040 & 0.108 & 0.926 & 0.856 & 0.769 & \textbf{0.311} & 0.041 & 9660.060 \\
  BQ) (tuned)      & $\pm$0.01 & $\pm$0.02 & $\pm$0.01 & $\pm$0.01 & $\pm$0.01 & $\pm$0.02 & $\pm$0.02 & $\pm$0.04 & \textbf{$\pm$0.03} & $\pm$0.01 & $\pm$514.65 \\
  \hline
  XGB              & 0.842 & 0.748 & 0.759 & 0.268 & 0.367 & 0.920 & 0.844 & 0.739 & 0.432 & 0.066 & 14444.307 \\
  (tuned)          & $\pm$0.02 & $\pm$0.02 & $\pm$0.02 & $\pm$0.03 & $\pm$0.03 & $\pm$0.02 & $\pm$0.02 & $\pm$0.04 & $\pm$0.08 & $\pm$0.03 & $\pm$11.99 \\
  \hline
  CatBoost         & 0.832 & 0.776 & 0.790 & 0.186 & 0.285 & 0.920 & 0.844 & 0.741 & 0.408 & 0.057 & 14437.103 \\
  (tuned)          & $\pm$0.02 & $\pm$0.02 & $\pm$0.02 & $\pm$0.02 & $\pm$0.03 & $\pm$0.02 & $\pm$0.02 & $\pm$0.04 & $\pm$0.06 & $\pm$0.02 & $\pm$4.79 \\
  \hline
  LightGBM         & 0.781 & 0.720 & 0.767 & 0.252 & 0.361 & 0.915 & 0.841 & 0.741 & 0.443 & 0.063 & 14410.417 \\
  (tuned)          & $\pm$0.02 & $\pm$0.03 & $\pm$0.02 & $\pm$0.03 & $\pm$0.04 & $\pm$0.02 & $\pm$0.02 & $\pm$0.04 & $\pm$0.11 & $\pm$0.02 & $\pm$1.37 \\
  \hline
  CatBoost         & 0.761 & 0.731 & 0.783 & 0.170 & 0.249 & 0.913 & 0.839 & 0.748 & 0.404 & 0.053 & 5.874 \\
  (default)        & $\pm$0.02 & $\pm$0.02 & $\pm$0.02 & $\pm$0.02 & $\pm$0.02 & $\pm$0.02 & $\pm$0.02 & $\pm$0.04 & $\pm$0.04 & $\pm$0.01 & $\pm$0.74 \\
  \hline
  Random Forest    & 0.727 & 0.650 & 0.644 & 0.376 & 0.462 & 0.913 & 0.834 & 0.716 & 0.386 & 0.074 & 14404.904 \\
  (tuned)          & $\pm$0.02 & $\pm$0.03 & $\pm$0.03 & $\pm$0.04 & $\pm$0.03 & $\pm$0.02 & $\pm$0.02 & $\pm$0.05 & $\pm$0.07 & $\pm$0.02 & $\pm$0.15 \\
  \hline
  LightGBM         & 0.693 & 0.684 & 0.747 & 0.307 & 0.407 & 0.908 & 0.836 & 0.745 & 0.461 & 0.068 & 0.583 \\
  (default)        & $\pm$0.03 & $\pm$0.03 & $\pm$0.03 & $\pm$0.03 & $\pm$0.04 & $\pm$0.02 & $\pm$0.02 & $\pm$0.04 & $\pm$0.06 & $\pm$0.02 & $\pm$0.06 \\
  \hline
  XGB              & 0.665 & 0.643 & 0.725 & 0.330 & 0.533 & 0.906 & 0.834 & 0.743 & 0.468 & 0.079 & 0.814 \\
  (default)        & $\pm$0.03 & $\pm$0.03 & $\pm$0.03 & $\pm$0.03 & $\pm$0.04 & $\pm$0.02 & $\pm$0.02 & $\pm$0.04 & $\pm$0.06 & $\pm$0.02 & $\pm$0.09 \\
  \hline
  Random Forest    & 0.640 & 0.633 & 0.672 & 0.553 & 0.425 & 0.907 & 0.833 & 0.727 & 0.432 & 0.073 & 0.488 \\
  (default)        & $\pm$0.04 & $\pm$0.03 & $\pm$0.03 & $\pm$0.04 & $\pm$0.03 & $\pm$0.02 & $\pm$0.02 & $\pm$0.04 & $\pm$0.19 & $\pm$0.02 & $\pm$0.03 \\
  \hline
  SVM              & 0.571 & 0.531 & 0.537 & 0.292 & 0.169 & 0.887 & 0.810 & 0.680 & 0.455 & 0.044 & 14412.047 \\
  (tuned)          & $\pm$0.03 & $\pm$0.03 & $\pm$0.03 & $\pm$0.03 & $\pm$0.02 & $\pm$0.02 & $\pm$0.02 & $\pm$0.04 & $\pm$0.04 & $\pm$0.01 & $\pm$3.05 \\
  \hline
  MLP              & 0.512 & 0.442 & 0.480 & 0.345 & 0.294 & 0.883 & 0.802 & 0.664 & 0.493 & 0.058 & 2.133 \\
  (default)        & $\pm$0.03 & $\pm$0.03 & $\pm$0.03 & $\pm$0.03 & $\pm$0.03 & $\pm$0.02 & $\pm$0.02 & $\pm$0.05 & $\pm$0.05 & $\pm$0.02 & $\pm$0.19 \\
  \hline
  MLP (sklearn)    & 0.458 & 0.411 & 0.448 & 0.432 & 0.306 & 0.877 & 0.800 & 0.653 & 0.764 & 0.059 & 14408.730 \\
  (tuned)          & $\pm$0.03 & $\pm$0.03 & $\pm$0.03 & $\pm$0.04 & $\pm$0.03 & $\pm$0.02 & $\pm$0.02 & $\pm$0.06 & $\pm$0.65 & $\pm$0.02 & $\pm$0.34 \\
  \hline
  Log.\ Regr.      & 0.401 & 0.354 & 0.391 & 0.386 & 0.241 & 0.874 & 0.789 & 0.637 & \textit{inf} & 0.049 & 14406.416 \\
  (tuned)          & $\pm$0.04 & $\pm$0.03 & $\pm$0.04 & $\pm$0.04 & $\pm$0.03 & $\pm$0.02 & $\pm$0.02 & $\pm$0.04 & $\pm$0.03 & $\pm$0.02 & $\pm$0.47 \\
  \hline
  SVM              & 0.388 & 0.406 & 0.430 & 0.357 & 0.202 & 0.872 & 0.794 & 0.672 & 0.482 & 0.046 & 2.887 \\
  (default)        & $\pm$0.04 & $\pm$0.03 & $\pm$0.03 & $\pm$0.04 & $\pm$0.02 & $\pm$0.02 & $\pm$0.02 & $\pm$0.04 & $\pm$0.03 & $\pm$0.01 & $\pm$0.60 \\
  \hline
  Log.\ Regr.      & 0.209 & 0.185 & 0.186 & 0.483 & 0.348 & 0.857 & 0.778 & 0.600 & 0.529 & 0.062 & 0.609 \\
  (default)        & $\pm$0.03 & $\pm$0.03 & $\pm$0.03 & $\pm$0.04 & $\pm$0.04 & $\pm$0.02 & $\pm$0.02 & $\pm$0.04 & $\pm$0.03 & $\pm$0.02 & $\pm$0.10 \\
  \hline
  \end{tabular}
  }}
\endgroup



\end{document}